\documentclass[sigconf]{acmart}
\renewcommand\footnotetextcopyrightpermission[1]{} 

\usepackage{listings} 
\usepackage{graphicx}
\usepackage{subfigure}
\usepackage{booktabs}
\usepackage{todonotes}
\usepackage{amsmath,amsfonts,amsthm}
\usepackage{textcomp}
\usepackage{xcolor}
\usepackage{multirow}
\usepackage{enumitem}
\usepackage{url}
\usepackage{soul} 

\usepackage[ruled, linesnumbered]{algorithm2e}
\usepackage{bm}
\usepackage{diagbox}

\AtBeginDocument{%
  \providecommand\BibTeX{{%
    \normalfont B\kern-0.5em{\scshape i\kern-0.25em b}\kern-0.8em\TeX}}}

\def\numx#1e#2{{#1}\mathrm{e}{#2}}

\soulregister{\ref}7
\soulregister{\cite}7

\begin{document}


\title{Deep Intention-Aware Network for Click-Through Rate Prediction}

\author{Yaxian Xia}
\email{yaxian.xyx@alibaba-inc.com}
\affiliation{%
  \country{}
  \institution{Alibaba Group}
}
\author{Yi Cao}
\email{dylan.cy@alibaba-inc.com}
\affiliation{%
  \country{}
  \institution{Alibaba Group}
}

\author{Sihao Hu}
\authornote{Sihao Hu is the corresponding author.}
\email{sihaohu@gatech.edu}
\affiliation{%
  \country{}
  \institution{Georgia Institute of Technology}
}

\author{Tong Liu}
\email{yingmu@taobao.com}
\affiliation{%
  \country{}
  \institution{Alibaba Group}
}

\author{Lingling Lu}
\email{lulingling@zju.edu.cn}
\affiliation{%
  \country{}
  \institution{Zhejiang University}
}

\begin{abstract}

E-commerce platforms provide entrances for customers to enter mini-apps that can meet their specific shopping requirements. Trigger items displayed on entrance icons can attract more entering. However, conventional Click-Through-Rate (CTR) prediction models, which ignore user instant interest in trigger item, fail to be applied to the new recommendation scenario dubbed Trigger-Induced Recommendation in Mini-Apps (TIRA). Moreover, due to the high stickiness of customers to mini-apps, we argue that existing trigger-based methods that over-emphasize the importance of trigger items, are undesired for TIRA, since a large portion of customer entries are because of their routine shopping habits instead of triggers. We identify that the key to TIRA is to extract customers' personalized entering intention and weigh the impact of triggers based on this intention. To achieve this goal, we convert CTR prediction for TIRA into a separate estimation form, and present Deep Intention-Aware Network (DIAN) with three key elements: 1) \textit{Intent Net} that estimates user's entering intention, \textit{i.e.}, whether he/she is affected by the trigger or by the habits; 2) \textit{Trigger-Aware Net} and 3) \textit{Trigger-Free Net} that estimate CTRs given user's intention is to the trigger-item and the mini-app respectively. Following a joint learning way, DIAN can both accurately predict user intention and dynamically balance the results of trigger-free and trigger-based recommendations based on the estimated intention.
Experiments show that DIAN advances state-of-the-art performance in a large real-world dataset, and brings a \textbf{9.39}\% lift of online Item Page View and \textbf{4.74}\% CTR for Juhuasuan, a famous mini-app of Taobao.

\end{abstract}


\maketitle

\section{Introduction}

Mini-app, a new marketing tool that can meet specific shopping requirements for customers, has been widely adopted in e-commerce platforms and contributes to billions of daily page views. Take Taobao for instance, Juhuasuan is a mini-app that only sells brand-discounted products, and Little Black Box is another mini-app that sells new products only. Another mini-app example is the Flash Deals of Shoppee, where customers can snap up low-priced products within time limits.



\begin{figure}[tbp!]
\centering
\includegraphics[width=7cm]{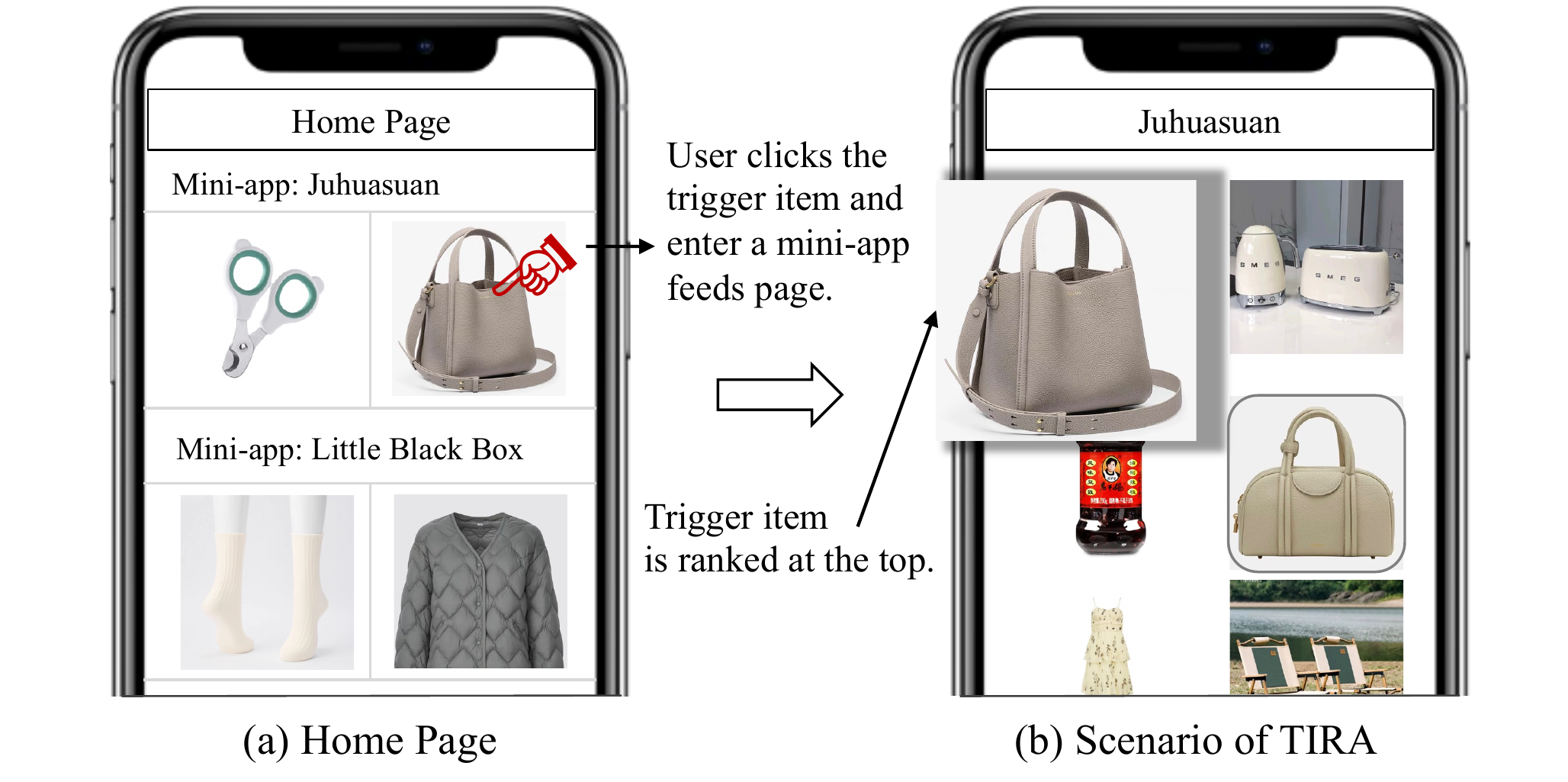}
\caption{Trigger-Induced Recommendation in Mini-Apps}
\label{fig:tira}
\end{figure}  

For e-commerce Apps, fixed entrances are always provided for customers to click and enter mini-apps. To further attract their attention, one or two trigger items are displayed on each entrance icon, which are personally recommended based on a user's historical behaviors, as shown in Figure~\ref{fig:tira}(a). Once a user clicks the entrance icon, he/she will enter a mini-app feeds page where the clicked trigger item is ranked at the top, and other recommended items follow up, as shown in Figure~\ref{fig:tira}(b). 
In this work, we focus on recommending follow-up items in the mini-app (feeds), given that a user has already clicked trigger item and entered a mini-app. We define this recommendation task as Trigger-Induced Recommendation in mini-Apps (TIRA).

The existence of trigger-item invalidates many popular CTR models like DIN\cite{DIN}, DIEN\cite{zhou2019deep}, SIM\cite{pi2020search} and GIFT\cite{cao2022gift}, because they cannot take into consideration the effect of trigger item that can reveal the instant interest when a user enters the mini-app.

\begin{figure}[tbp!]
\centering
\includegraphics[width=7cm]{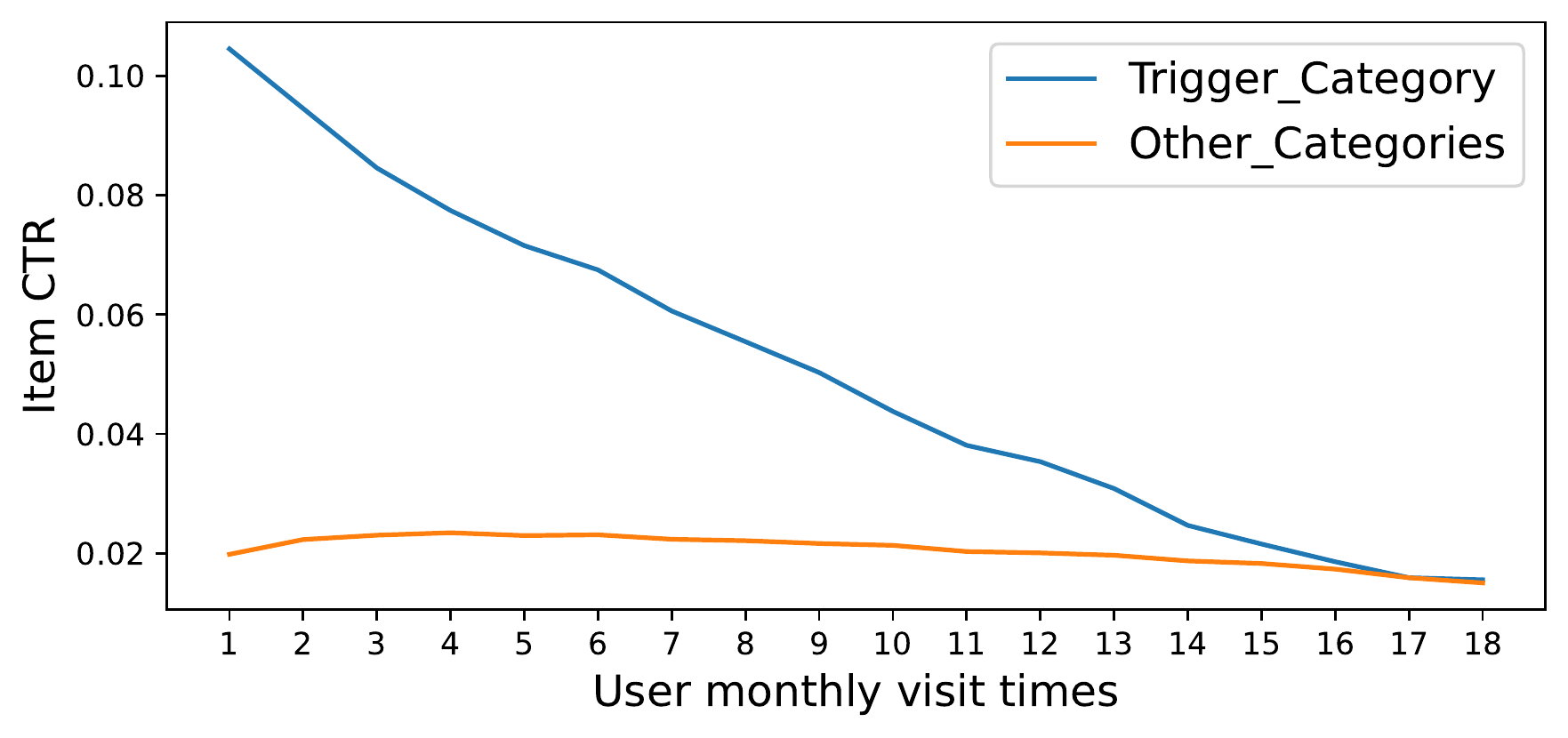}
\caption{Average CTR of items from trigger category and other categories \textit{w.r.t.} user monthly visit times in Juhuasuan.}
\label{fig:data_analysis}
\end{figure}

Moreover, the characteristic of mini-app prevents trigger-based methods\cite{DBLP:conf/kdd/LinWMZWJW22,www_dihn,wsdm21_RelevantRecommendation,10.1145/3437963.3441733} being applied to TIRA: Regular customers of mini-apps exhibit a strong stickiness to specific mini-apps, indicating that their entering behaviors are not likely affected by triggers but by their routine shopping habits. This observation can be proved by Figure~\ref{fig:data_analysis}: users who visit a mini-app more than 15 times a month have almost the same preference for the items from the trigger category and other categories. In this case, trigger-based methods that over-emphasize the importance of triggers and thus recommend follow-up items with high relevance to triggers are undesired. For regular users, recommendation results should exhibit diversity to meet their multiple interests to the mini-app.

Therefore, the most important objective for TIRA is to strike a balance between trigger-free and trigger-based recommendations via identifying customers' personalized entering intention. To achieve this goal, we present Deep Intention-Aware Network (DIAN) that explicitly recognizes user entering intention by an auxiliary task, and dynamically adjusts recommendation strategy. The whole DIAN consists of three key elements, dubbed Intention Net, Trigger-Aware Net and Trigger-Free Net. The Intention Net is designed to estimate a user's entering intention based on the his/her historical preference toward the trigger item or the mini-app. To improve the accuracy of intention score estimated by the Intention Net, we introduce a posterior labeling strategy and an auxiliary training task, making our estimation can achieve 91.3\% accuracy.

Trigger-Aware Net is proposed to estimate a conditional CTR by assuming that a user is attracted by the trigger item only. We elaborately design the model architecture to make it fully exploit the user's instant interest revealed by the trigger item. Conversely, Trigger-Free Net is proposed to estimate a conditional CTR by assuming a user's entering is only led by his/her routine habits and preference to the mini-app. Following a joint-learning paradigm, DIAN fuses the outputs of Trigger-Aware Net and Trigger-Free Net based on the estimated user intention score, making it dynamically adjust recommendation result according to customers with different entering intention.

Offline experiments show that DIAN advances state-of-the-art performance in a large real-world dataset. Online A/B testing suggests that DIAN can bring a \textbf{9.39}\% lift of online Item Page View and \textbf{4.74}\% CTR for Juhuasuan, a famous mini-app of Taobao. Currently, DIAN has been deployed on Taobao App since Sep. 21, 2022, serving for the full volume traffic of Juhuasuan.
 


\noindent \textbf{Contributions.} To summarize, the contributions are as follows:
\begin{itemize}[leftmargin=10 pt]
  \item We define a new task of Trigger-Induced Recommendation in Mini-Apps (TIRA) and identify problems of existing trigger-free and trigger-based CTR prediction models for TIRA.

  \item We propose DIAN equipped with novel architecture and effective intention estimation strategy, making it can accurately reveal user's personalized intention and dynamically adjust the recommendation result.

  \item DIAN significantly advances the state-of-the-art on an large industrial dataset. Online A/B testing indicates DIAN can bring huge benefit to the real-world application.
  
\end{itemize}

\section{Proposed Method}

\subsection{TIRA Problem Formulation}
\label{sec:TIRA}
CTR prediction task aims to estimate a click-through probability $P(Click|Target)$, given a user be recommended a target item\cite{jicai_dsin,www2020_ctr,www2021_ctr}. For TIRA, we take into consideration the impact of users' personalized entering intention to CTR estimation. We use $Intent=0$ and $Intent=1$ to indicate whether a user's entering intention is the mini-app or the trigger-item. Therefore, user intention can be represented as a probability  distribution ($P(Intent=0)$, $P(Intent=1)$), and $P(Intent=0) + P(Intent=1) = 1$ since there are only two reasons of entering the mini-app.

In this case, CTR estimation can be converted to another form by separately estimating 1) User intention, 2) CTR given user's intention to the trigger item, and 3) CTR given user's intention to the mini-app:
\begin{small}
\begin{equation}
\label{eq:total_prob}
 \begin{split}
P(Click\mid Target) &= P(Click\mid Target, Intent=0) \ast P(Intent=0) \\ &+P(Click\mid Target,Intent=1)\ast P(Intent=1)
\end{split}
\end{equation}
\end{small}
The advantage of this separate estimation way is that we can design individual neural networks to model user's interests to the target item given two different intentions, and leverage the estimated user intention score to balance two different types of user interests. Obviously, existing trigger-free and trigger-based methods cannot take both intentions into account.

To model above-mentioned three elements, we propose Deep Intention-Aware Network (DIAN), which is equipped with three key sub-nets: 1) Intention Net that aims to estimate user entering intention; 2) Trigger-Aware Net that aims to estimate CTR given user intention to the trigger-item; 3) Trigger-Free Net that aims to estimate CTR given user intention to the mini-app. We introduce them in detail one-by-one.

\begin{figure*}[tbp!]
\centering
\includegraphics[width=15.5cm]{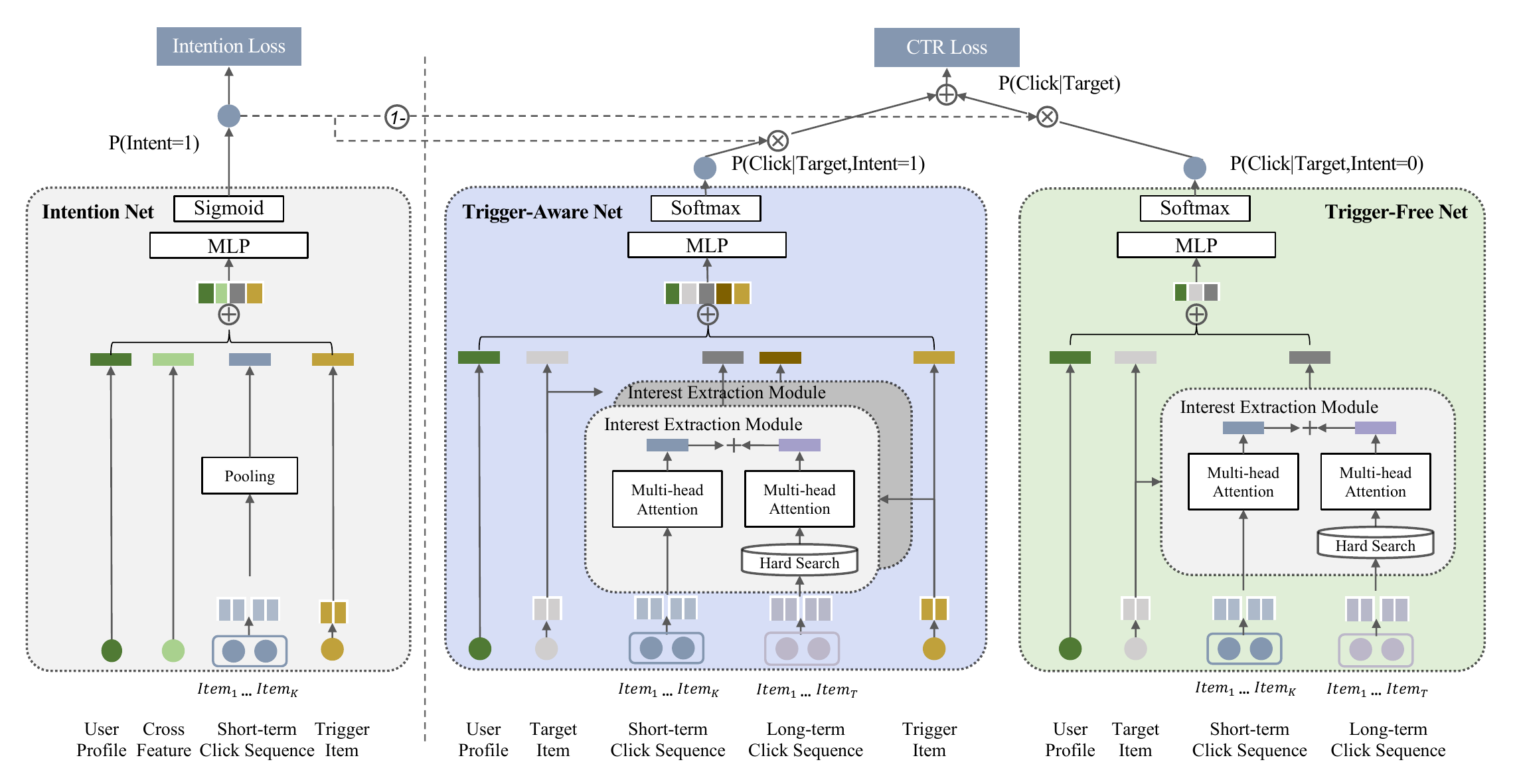}
\caption{DIAN consists of three sub-nets: Intention Net, Trigger-Aware Net and Trigger-Free Net.}
\label{fig:framework}
\vspace{-0.1cm}
\end{figure*}

\subsection{Intention Net}

We start with introducing the strategy of identifying user entering intention in the training set.

\noindent \textbf{User Intention Labeling:} After entering the mini-app page, if a user still clicks on the trigger item or items with the same leaf-category (a relative strict category definition in Taobao), we label his/her entering intention to the trigger item ($Intent=1$). Otherwise, we label the entering intention to the mini app ($Intent=0$), because the user doesn't interested in the trigger item or similar items. It should be noted that although we can identify user entering intention, this posterior labeling strategy cannot be use for online serving because the recommendation happens at the moment the user enters the mini-app. Therefore, we need to estimate the intention score by Intention Net.

\noindent \textbf{User Intention Estimation:} To estimate the personalized intention score for each user, Intention Net takes four parts as the input: 1) User profile features like user\_id, age, occupation, \textit{etc.} 2) User short-term click item sequence, which reflects his/her historical interest. 3) User-mini-app cross features that measure the user's preference to the mini-app, including his/her monthly visit times or average stay duration. 4) Trigger item features like item\_id, category\_id, \textit{etc.} We use widely adopted embedding technique to project them into embedding vectors as $H_{user}$, $H_{sseq}$, $H_{crs}$, and $H_{tri}$, respectively. The sequence vector $H_{sseq}$ is the average pooling result of embeddings of items in the click sequence.


We concatenate four embedding vectors together and pass them through a MLP layer, which can be formalized as:
\begin{equation}
\label{eq:intention_net}
\hat{y}_{Intent} = \operatorname{Sigmoid} \left(\operatorname{MLP}(H_{user} \oplus H_{sseq}  \oplus H_{crs} \oplus H_{tri})\right)
\end{equation}
where $\hat{y}$ represents the user's intention probability $P(intent)$. $\oplus$ is the concatenate operator.

\subsection{Trigger-Aware Net}
Trigger-Aware Net (TAN) is proposed to estimate the CTR of the target item by assuming a user's entry is due to the trigger item. In this case, the trigger item is critical for recommending follow-up items. Therefore, we design TAN to highlight the impact of triggers and to fully exploit the instant interest revealed by triggers.




The input of TAN consists of five parts: 1) User profile features. 2) User short-term click sequence. 3) User long-term click sequence. 4) Trigger item features. 5) Target item features. Similarly as in Intention Net, we adopt embedding techniques to project these original features into embedding vectors and denote them as $E_{user}$, $\boldsymbol{E}_{sseq}$, $\boldsymbol{E}_{lseq}$, $E_{tri}$ and $E_{tar}$. $\boldsymbol{E}_{sseq}=[E_{item}^{s1},\cdots,E_{item}^{sK}]$ and $\boldsymbol{E}_{lseq}=[E_{item}^{l1},\cdots,E_{item}^{lT}]$.




\noindent \textbf{Dual-layer Interest Extraction:} We extract user interest information contained within short-term and long-term click sequences separately according to the trigger item and target item, to prevent them from disturbing each other when they are diverse. For trigger item and target item, we generate user interest representation $h_{tri}$ and $h_{tar}$ respectively as follows:
\begin{equation}
    H_{tri} = \operatorname{InterestExtract_{tri}}(E_{tri},\boldsymbol{E}_{sseq}, \boldsymbol{E}_{lseq})
\label{eq:interest_eq_tri}
\end{equation}
\begin{equation}
    H_{tar} = \operatorname{InterestExtract_{tar}}(E_{tar},\boldsymbol{E}_{sseq}, \boldsymbol{E}_{lseq})
\label{eq:interest_eq_tar}
\end{equation}
Take Eq.~\ref{eq:interest_eq_tri} as an example, we introduce one layer of Interest Extraction (IE) in detail. An IE layer consists of short-term IE and long-term IE modules, where short-term IE module tackles click sequence of the past 14 days and long-term IE tackles click sequence of the past 180 days.







For short-term IE, we apply multi-head\cite{transformer} target attention to retrieve relevant information within the short-term click sequence with the trigger item embedding $E_{tri}$, to generate a user short-term interest embedding $H_{tri}^{s}$:
\begin{small}
\begin{equation}
\label{eq:ss_tri}
 \begin{split}
{head}_{i} &= \boldsymbol{W}^V_i \boldsymbol{E}_{sseq} \cdot \operatorname{Softmax}\left(\boldsymbol{W}^Q_i E_{tri} \odot \boldsymbol{W}^K_i \cdot \boldsymbol{E}_{sseq}\right)
 \\
H_{tri}^{s} &=\left[{head}_{1}\oplus {head}_{2} \cdots\oplus{head}_{n}\right] 
\end{split}
\end{equation}
\end{small}
where $\odot$ is broadcasting dot-product operator, projection matrices for each head $\boldsymbol{W}_i^{Q}$, $\boldsymbol{W}_i^{K} \in \mathbb{R}^{d \times d / n} $ are learnable parameters and $d$ is the dimension of hidden space.





User's long-term interest can bring more diverse recommendation results compared to short-term interest. To generate the long-term interest representation $h_{tri}^{l}$ toward the trigger item, we adopt the search-based IE method proposed by SIM\cite{pi2020search}. Specifically, it hard-searches the most recent $K$ items, ${S}_{lseq}$, belonging to the same leaf category of the trigger item, which can largely reduce the length of long-term click sequence.
Similar to short-term IE, we apply another multi-head target attention function to softly retrieve relevant information within a filtered click sequence by the trigger item:
\begin{small}
\begin{equation}
\label{eq:ls_tri}
 \begin{split}
{head}_{j} &= \boldsymbol{W}^V_j \boldsymbol{S}_{lseq} \cdot \operatorname{Softmax}\left(\boldsymbol{W}^Q_j E_{tri} \odot \boldsymbol{W}^K_j \cdot \boldsymbol{S}_{lseq}\right)
 \\
H_{tri}^{l} &=\left[{head}_{1}\oplus {head}_{2} \cdots\oplus{head}_{n}\right] 
\end{split}
\end{equation}
\end{small}

It should be noted that projection matrices of long-term IE layer are not shared with short-term IE layer. The subscript $lseq$ and $sseq$ are omitted for the sake of simplicity. By adding short-term and long-term user interest embeddings, we obtain the user interest representation toward the trigger item, \textit{i.e.}, $H_{tri}=H_{tri}^{sseq} + H_{tri}^{lseq}$.





For the target item, just following the same calculation, we can obtain the user interest representation $H_{tar}$. Finally, we concatenate $E_{user}$, $E_{tri}$, $E_{tar}$, $H_{tri}$ and $H_{tar}$ together and pass them through a MLP layer:
\begin{equation}
\label{eq:intention_net}
\hat{y}_{tan} = \operatorname{Sigmoid} \left(\operatorname{MLP}(E_{user} \oplus E_{tri} \oplus E_{tar} \oplus H_{tri} \oplus H_{tar})\right)
\end{equation}
where $\hat{y}_{tan}$ represents the CTR to target item, given user's entering intention is to trigger-item, \textit{i.e.}, $P(Click\mid Target,Intent=1)$.



\subsection{Trigger-Free Net}
Trigger-Free Net (TFN) aims to estimate CTR given user's intention is to the mini-app. As shown in Figure~\ref{fig:framework}, we adopt a similar architecture of TAN but exclude trigger item features from the input and remove the interest extraction layer for trigger-item, making it degrade into SIM\cite{pi2020search}. To reduce the size of parameters, we force TFN share the same embedding layer with Trigger-Aware Net, since embedding layer accounts the most large portion of parameters in CTR models.

Following the same way as in Eq.\ref{eq:interest_eq_tar}, we generate user interest representation $H_{tar}^{'}$ to the target item. We concatenate three embeddings together and pass them through a MLP layer:
\begin{equation}
\label{eq:TFN}
\hat{y}_{tfn} = \operatorname{Sigmoid} \left(\operatorname{MLP}(E_{user} \oplus E_{tar} \oplus H_{tar}^{'})\right)
\end{equation}
where $\hat{y}_{tfn}$ represents the CTR to target item, given user's entering intention is to mini-app, \textit{i.e.}, $P(Click\mid Target,Intent=0)$.

\subsection{Multi-Task Training}

Given estimated user intention $\hat{y}_{int}$, and CTRs given user's different intention $\hat{y}_{tan}$ and $\hat{y}_{tfn}$, we can estimate the final CTR as illustrated in Section~\ref{sec:TIRA} (Eq.~\ref{eq:total_prob}):
\begin{small}
\begin{equation}
\hat{y} = \hat{y}_{int}*\hat{y}_{tan} + (1-\hat{y}_{int}) * \hat{y}_{tfn}
\end{equation}
\end{small}
The final objective function consists of CTR loss and intention loss:
\begin{small}
\begin{equation}
\label{eq:final_loss}
\begin{split}
L=-\frac{1}{|\mathcal{D}|} \sum_{(\hat{y}, y) \in \mathcal{D}}\Big( y \log \hat{y}+(1-y) \log (1-\hat{y}) \\
+ \alpha \big (y_{int} \log \hat{y}_{int}+(1-y_{int}) \log (1-\hat{y}_{int}) \big) \Big)
\end{split}
\end{equation}
\end{small}
where $\mathcal{D}$ is the training set, $y\in\{0,1\}$ is the label for CTR prediction task, and $y_{int}\in\{0,1\}$ is the label for user intention estimation task, \textit{i.e.}, it denotes whether a user's entering intention is the mini-app or trigger item. $\alpha$ is hyper-parameter to balance CTR loss and intention loss, which is set to 0.1 in our experiments.

\section{Experiments}

\begin{table}
\small
\caption{CTR prediction comparison on the testing set}
\begin{center}
\begin{tabular}{p{3.5cm}<{\centering}p{2.5cm}<{\centering}}
\toprule
{\textbf{Method}} & {\textbf{AUC}}\\
\midrule
Wide \& Deep & 0.767  \\
DIN & 0.791  \\
SIM (TFN) & 0.800 \\
\midrule
DIHN & 0.805 \\
TAN & 0.806 \\
\midrule
DIAN &\textbf{0.812}\\
DIAN (remove intention loss) & 0.808 \\
DIAN (remove Intention Net) & 0.806  \\
\bottomrule
\end{tabular}
\end{center}
\label{tab:compare}
\end{table}

\begin{table}
\caption{Intention estimation results on the testing set}
\small
\begin{center}
\begin{tabular}{p{2cm}<{\centering}p{1cm}<{\centering}p{1cm}<{\centering}p{1cm}<{\centering}p{1cm}<{\centering}}
\toprule
{\textbf{Method}} & {\textbf{Accuracy}} & {\textbf{AUC}}\\
\midrule
Intention Net & 91.3\% & 0.943 \\
\bottomrule
\end{tabular}
\end{center}
\label{tab:intention}
\end{table}

\subsection{Experimental Setup}

\noindent \textbf{Dataset:} we collect a dataset from Juhuasuan in Taobao, which is a famous mini-app that sells brand-discounted products. We first sample a training set from user impression and click logs from Aug. 1, to Sep. 1, 2022, and then sample the testing set in Sep. 2, 2022, making it strictly simulates online serving. The training set has 10,430,212,219 samples and testing set has 352,959,879 samples, with in total 173,406,246 users and 6,500,833 items involved.

\noindent \textbf{Baselines:} We compare DIAN against two types of baselines that are widely used in the industry: 1) Trigger-free methods: Wide \& Deep\cite{WideDeep}, DIN\cite{DIN}, SIM\cite{pi2020search}; 2) Trigger-based methods: DIHN\cite{www_dihn} and our Trigger-Aware Net (TAN).



\noindent \textbf{Implementation:} For all the competitors, dimensions of hidden layers are set to $1024 \times 512 \times 256$. Among sequential models, short-term click sequence is used for DIN and both short- and long-term sequences are used for SIM(TFN), DIHN, TAN and DIAN. Adam optimizer is adopted with learning rate of 0.01. For all attention-based modules, the number of attention heads is set to 4.

\subsection{Offline Evaluation}

\textbf{Performance Comparison:} Table \ref{tab:compare} shows the comparison results of CTR prediction task on the testing set, where we adopt Area Under Curve (AUC) that widely used for industrial recommendation as the measurement metric. 
The first observation is that trigger-based methods outperforms trigger-free methods, indicating that trigger item that reveals user instant interest is important for TIRA task; Second, among trigger-based methods, DIHN works not very well because it dedicate to improve the relevance between recommendation results and the trigger item, which is undesired for those routine users of mini-app.
The last observation is that the proposed DIAN network outperforms all the competitors. We contribute its superior performance to it explicitly identifies user's entering intention thus strike a perfect balance between trigger-free and trigger-based recommendations, making it suitable for TIRA.



\noindent \textbf{Ablation Study:} We verify effectiveness of each element of DIAN in Table~\ref{tab:compare}. The first observation is that TFN and TAN can harmoniously work together and their performance can be improved by the Intention Net. This is consistent with the second observation that when we remove the Intention Net and averages outputs of TAN and TFN, AUC value decreases to 0.806.
Finally, if we remove intention loss to make DIAN itself balance the outputs of TAN and TFN, the performance also decreases a lot, indicating that it is hard to identify user entering intention within just impression-click samples. The proposed labeling strategy and auxiliary intention estimation task are essential to truly identify user's intention.

\noindent \textbf{User Intention Estimation:} Table \ref{tab:intention} summarizes the results of intention estimation task, where AUC value reaches 0.943 and accuracy reaches 91.3\%, indicating that the proposed Intention Net is effective to identify user's entering intention.



\begin{table}
\small
\begin{center}
\caption{Online A/B testing results}
\begin{tabular}{p{1.5cm}<{\centering}p{1cm}<{\centering}p{1cm}<{\centering}p{1cm}<{\centering}}
\toprule
{\textbf{Method}} & {\textbf{IPV}} & {\textbf{PV}} & {\textbf{CTR}}\\
\midrule
SIM & 0.555 & 23.93 & 2.32\% \\
DIAN & 0.607 & 24.91 & 2.43\%\\
\midrule 
Lift rate & +9.39\% & +4.10\% & +4.74\%\\
\bottomrule
\end{tabular}
\label{tab:abtest}
\end{center}
\end{table}
\subsection{Online A/B Testing}

We implemented and deployed the proposed DIAN for Juhusuan mini-app in Taobao App. From Sep. 16 to Oct. 15, 2022, a strict online A/B test is conducted to compare DIAN against the previous online model SIM. Table~\ref{tab:abtest} shows the comparison results \textit{w.r.t.} Page View (PV), Item Page View (IPV) and CTR, where PV and IPV are the average numbers of items that users browse and click after they enter the mini-app. 
It is obvious that DIAN outperforms SIM with a large margin: DIAN contributes up to $4.74\%$ lift for CTR, $4.10\%$ lift for PV and $9.39\%$ lift for IPV, which indicates that users not only browse more but also more likely to click items recommended by DIAN. 


\section{Conclusion}
In this paper, we define a new task dubbed Trigger-Induced Recommendation in mini-App (TIRA) and identify the key to TIRA is to accurately estimate user's intention when they enter the mini-app. A novel Deep Intention-Aware Network is proposed to achieve this goal. Experiments on a large industrial dataset and online A/B testing verify its effectiveness for the real-world application. Currently, DIAN has been deployed on Taobao App to serve for the full volume traffic of Juhusuan.

\bibliographystyle{abbrv}
\bibliography{DIAN}

\begin{thebibliography}{10}

\bibitem{cao2022gift}
Y.~Cao, S.~Hu, Y.~Gong, Z.~Li, Y.~Yang, Q.~Liu, and S.~Ji.
\newblock Gift: Graph-guided feature transfer for cold-start video
  click-through rate prediction.
\newblock In {\em Proceedings of the 31st ACM International Conference on
  Information \& Knowledge Management}, pages 2964--2973, 2022.

\bibitem{WideDeep}
H.-T. Cheng, L.~Koc, J.~Harmsen, T.~Shaked, et~al.
\newblock Wide \& deep learning for recommender systems.
\newblock In {\em Proceedings of the 1st workshop on deep learning for
  recommender systems}, 2016.

\bibitem{jicai_dsin}
Y.~Feng, F.~Lv, W.~Shen, M.~Wang, F.~Sun, Y.~Zhu, and K.~Yang.
\newblock Deep session interest network for click-through rate prediction.
\newblock In S.~Kraus, editor, {\em Proceedings of the Twenty-Eighth
  International Joint Conference on Artificial Intelligence, {IJCAI} 2019,
  Macao, China, August 10-16, 2019}, pages 2301--2307. ijcai.org, 2019.

\bibitem{www2020_ctr}
X.~Li, C.~Wang, J.~Tan, X.~Zeng, D.~Ou, and B.~Zheng.
\newblock Adversarial multimodal representation learning for click-through rate
  prediction.
\newblock In Y.~Huang, I.~King, T.~Liu, and M.~van Steen, editors, {\em {WWW}
  '20: The Web Conference 2020, Taipei, Taiwan, April 20-24, 2020}, pages
  827--836. {ACM} / {IW3C2}, 2020.

\bibitem{DBLP:conf/kdd/LinWMZWJW22}
Z.~Lin, H.~Wang, J.~Mao, W.~X. Zhao, C.~Wang, P.~Jiang, and J.~Wen.
\newblock Feature-aware diversified re-ranking with disentangled
  representations for relevant recommendation.
\newblock In A.~Zhang and H.~Rangwala, editors, {\em {KDD} '22: The 28th {ACM}
  {SIGKDD} Conference on Knowledge Discovery and Data Mining, Washington, DC,
  USA, August 14 - 18, 2022}, pages 3327--3335. {ACM}, 2022.

\bibitem{pi2020search}
Q.~Pi, G.~Zhou, Y.~Zhang, Z.~Wang, L.~Ren, Y.~Fan, X.~Zhu, and K.~Gai.
\newblock Search-based user interest modeling with lifelong sequential behavior
  data for click-through rate prediction.
\newblock In {\em Proceedings of the 29th ACM International Conference on
  Information \& Knowledge Management}, pages 2685--2692, 2020.

\bibitem{www_dihn}
Q.~Shen, H.~Wen, W.~Tao, J.~Zhang, F.~Lv, Z.~Chen, and Z.~Li.
\newblock Deep interest highlight network for click-through rate prediction in
  trigger-induced recommendation.
\newblock In F.~Laforest, R.~Troncy, E.~Simperl, D.~Agarwal, A.~Gionis,
  I.~Herman, and L.~M{\'{e}}dini, editors, {\em {WWW} '22: The {ACM} Web
  Conference 2022, Virtual Event, Lyon, France, April 25 - 29, 2022}, pages
  422--430. {ACM}, 2022.

\bibitem{transformer}
A.~Vaswani, N.~Shazeer, N.~Parmar, J.~Uszkoreit, et~al.
\newblock Attention is all you need.
\newblock {\em arXiv:1706.03762}, 2017.

\bibitem{wsdm21_RelevantRecommendation}
R.~Xie, R.~Wang, S.~Zhang, Z.~Yang, F.~Xia, and L.~Lin.
\newblock Real-time relevant recommendation suggestion.
\newblock In L.~Lewin{-}Eytan, D.~Carmel, E.~Yom{-}Tov, E.~Agichtein, and
  E.~Gabrilovich, editors, {\em {WSDM} '21, The Fourteenth {ACM} International
  Conference on Web Search and Data Mining, Virtual Event, Israel, March 8-12,
  2021}, pages 112--120. {ACM}, 2021.

\bibitem{10.1145/3437963.3441733}
R.~Xie, R.~Wang, S.~Zhang, Z.~Yang, F.~Xia, and L.~Lin.
\newblock Real-time relevant recommendation suggestion.
\newblock In {\em Proceedings of the 14th ACM International Conference on Web
  Search and Data Mining}, WSDM '21, page 112–120, New York, NY, USA, 2021.
  Association for Computing Machinery.

\bibitem{www2021_ctr}
S.~Yao, J.~Tan, X.~Chen, K.~Yang, R.~Xiao, H.~Deng, and X.~Wan.
\newblock Learning a product relevance model from click-through data in
  e-commerce.
\newblock In J.~Leskovec, M.~Grobelnik, M.~Najork, J.~Tang, and L.~Zia,
  editors, {\em {WWW} '21: The Web Conference 2021, Virtual Event / Ljubljana,
  Slovenia, April 19-23, 2021}, pages 2890--2899. {ACM} / {IW3C2}, 2021.

\bibitem{zhou2019deep}
G.~Zhou, N.~Mou, Y.~Fan, Q.~Pi, et~al.
\newblock Deep interest evolution network for click-through rate prediction.
\newblock In {\em AAAI}, 2019.

\bibitem{DIN}
G.~Zhou, X.~Zhu, C.~Song, Y.~Fan, et~al.
\newblock Deep interest network for click-through rate prediction.
\newblock In {\em SIGKDD}, 2018.

\end{thebibliography}
\clearpage

\appendix

\end{document}